# Co-Occurrence Patterns in the Voynich Manuscript

Torsten Timm
torsten.timm@kereti.de

**Abstract:** The Voynich Manuscript is a medieval book written in an unknown script. This paper studies the distribution of similarly spelled words in the Voynich Manuscript. It shows that the distribution of words within the manuscript is not compatible with natural languages.

## 1  Introduction

The Voynich Manuscript, a mysterious handwritten manuscript also referred to as the VMS, contains a text in an unknown script. The manuscript consists of 240 parchment pages. The parchment was carbon-dated to the early 15$^{th}$ century [see Stolte]. The script uses 20-30 different glyphs. The exact number is uncertain since it is unclear whether some of the glyphs are distinct characters or a ligature of two other characters. The text was written from left to right and apparently divided by spaces into word tokens. The modern history of the medieval manuscript starts in 1912, when the Polish-born bookseller Wilfried Voynich bought it from a Jesuit college in Italy. Since its discovery the manuscript has attracted the attention of numerous researchers. So far, however, nobody has been able to reveal the secret of the unknown script.

The Voynich Manuscript is unique since some features of it correspond to a natural language whereas others do not. The word frequency distribution behaves as expected according to Zipf's law. This law was proposed by George Kingsley Zipf in 1935 and makes predictions about the word frequencies for a natural language text. Moreover, it is possible to demonstrate that the word tokens depend on their context, a further feature similar to natural languages [Montemurro]. Untypical for a natural language is the weak word order [Reddy: p. 82]. Another untypical feature is the observation that the word length has a binomial distribution with an underrepresentation of short and long words.[1] This holds for both type and token lengths.[2]

A surprising feature of the VMS is that some elements are typical for a specific position within a line. For instance, in 86 % of cases the first word of a paragraph is highlighted by an additional gallow glyph (ℓ, ℓ, ℓ, ℓ) as the first symbol [Reddy: p. 80]. Moreover, the words in the first line of a paragraph are longer on average, their average length being 6.3 glyphs as opposed to 5.0 for all words. This is because, on average, words containing a gallow glyph are longer, and more words containing a gallow glyph occur in the initial lines of a paragraph [Timm: p. 29].
Within a paragraph the words at the beginning of a line more frequently start with an additional glyph ꝯ, o, ȣ or ꝫ than within a line (68 % and 50 % respectively) [Timm: p. 19]. In a statistical analysis Vogt demonstrates: "1. The first word with $i = 1$ of a line is *longer* than average, $\bar{l}_1 > \bar{l}$, 2. The second word with $i = 2$ is shorter, $\bar{l}_2 < \bar{l}$" [Vogt: p. 4]. In fact, the second word is shorter than the first word in 48% of the lines and longer in only 32 % [Timm: p. 19]. This means that to some extent the position of a word in a paragraph or line is encoded in the words.

---

[1] The length of the words is equally distributed around the arithmetical mean [see Stolfi]. The average word length is 5.5. Such a distribution is unusual because natural languages tend to make frequent use of short words. Typical for a natural language is therefore an asymmetric distribution.
[2] A word type describes a distinct word. A word token describes the occurrence of a word in the VMS text.

The connections between similarly spelled word types represent another remarkable feature. In 2013 Montemurro and Zanette analyzed the 100 most frequently used words within the VMS. They came to the conclusion that: "the words that are more strongly connected have an evident morphological similarity. Some of the words are linked by their prefixes, as in the case for the strongly connected pair chol-chor" (ᴄʜoʟ - ᴄʜoʀ) [Montemurro].

The pattern described by Montemurro and Zanette did not only hold for the 100 most frequent words. In fact, at least for the words occurring more than four times, it is possible to generate another word type from the word pool by replacing a glyph by a similar one, or by adding or deleting a glyph. It is surprising that in most cases all conceivable spelling permutations of a word exist. Moreover, it is possible to order the types by their similarities to build a multidimen­sional grid containing all word types which occur at least four times [see Timm: p. 66-82].[3]

| | | | |
|---|---|---|---|
| ᴄᴄȣꝯ (501) | ᴄᴄᴄȣꝯ ( 59) | ᴄ̈ᴄȣꝯ (426) | ᴄ̈ᴄᴄȣꝯ ( 84) |
| ᴄᴄꝯ (344) | ᴄᴄᴄꝯ (174) | ᴄ̈ᴄꝯ (283) | ᴄ̈ᴄᴄꝯ (144) |
| ᴄꝯ (155) | ᴄᴄȣꝯ (150) | ᴄ̈ꝯ (104) | ᴄ̈ᴄȣꝯ ( 46) |
| ᴄoȣꝯ ( 94) | ᴄᴄoȣꝯ ( 89) | ᴄ̈oȣꝯ ( 55) | ᴄ̈ᴄoȣꝯ ( 50) |
| ᴄᴄꝫꝯ ( 15) | ᴄᴄoꝫꝯ ( 5) | ᴄ̈oꝫꝯ ( 4) | ᴄ̈ᴄoꝫꝯ ( 1) |
| ᴄoꝯ ( 13) | ᴄᴄoꝯ ( 4) | ᴄ̈oꝯ ( 2) | ᴄ̈ᴄoꝯ ( 4) |

Table 1: sample of similar word types and their frequency of occurrence

Similarly spelled word types occur with predictable fre­quencies [Timm: p. 6]. They occur with comparable frequency, whereas types which contain less frequent glyphs or bigrams in most cases occur less frequently. For instance, if it is known that a word ccc8₉ ("chedy")[4] is frequent, it is possible to predict that the similar word cc8₉ ("shedy") is also frequently used, albeit less so, since c ("sh") occurs half as frequently as c ("ch").

---

[3] For words occurring less than four times transcription errors become important.

Similarly spelled types co-occur within the VMS. For each word it is nearly always possible to find a similarly spelled one within the same line or one or two lines above [Timm: p. 12]. To analyze this effect statistically it is possible to calculate the number of times identical or similar words occur near to each other for the whole VMS.

| | m-6 | m-5 | m-4 | m-3 | m-2 | m-1 | m | m+1 | m+2 | m+3 | m+4 | m+5 | m+6 |
|---|---|---|---|---|---|---|---|---|---|---|---|---|---|
| n-9 | 0.63% | 0.52% | 0.53% | 0.63% | 0.59% | 0.75% | 0.78% | 0.82% | 0.72% | 0.63% | 0.60% | 0.63% | 0.55% |
| n-8 | 0.52% | 0.55% | 0.69% | 0.70% | 0.74% | 0.68% | 0.67% | 0.70% | 0.74% | 0.77% | 0.68% | 0.62% | 0.64% |
| n-7 | 0.48% | 0.43% | 0.63% | 0.70% | 0.68% | 0.66% | 0.74% | 0.73% | 0.72% | 0.74% | 0.70% | 0.56% | 0.60% |
| n-6 | 0.52% | 0.53% | 0.69% | 0.69% | 0.75% | 0.67% | 0.76% | 0.71% | 0.84% | 0.75% | 0.69% | 0.69% | 0.57% |
| n-5 | 0.52% | 0.59% | 0.58% | 0.79% | 0.66% | 0.80% | 0.84% | 0.71% | 0.75% | 0.72% | 0.69% | 0.60% | 0.54% |
| n-4 | 0.67% | 0.61% | 0.66% | 0.78% | 0.78% | 0.81% | 0.83% | 0.66% | 0.71% | 0.69% | 0.69% | 0.63% | 0.44% |
| n-3 | 0.57% | 0.66% | 0.60% | 0.64% | 0.69% | 0.95% | 0.82% | 0.79% | 0.73% | 0.69% | 0.66% | 0.72% | 0.67% |
| n-2 | 0.56% | 0.71% | 0.73% | 0.86% | 0.87% | 0.89% | 0.93% | 0.96% | 0.92% | 0.87% | 0.77% | 0.67% | 0.59% |
| n-1 | 0.73% | 0.70% | 0.70% | 0.74% | 0.91% | 0.89% | 0.99% | 0.98% | 1.02% | 0.80% | 0.91% | 0.74% | 0.79% |
| n | 0.75% | 0.70% | 1.01% | 1.01% | 1.07% | 0.96% | 'x' | | | | | | |

**Table 2: The VMS – proportion of identical words appearing near to the word in writing position 'x'**

Table 2 shows the overall probabilities that a word is identical to a previously written word within the same context.[5] The 'x' stands for an imaginary writing position {n,m} in line 'n' in position 'm'. All probabilities are calculated by comparing each word with all previously written words within a range of eight lines. Each position is colored, whereby a darker color stands for a higher ratio of word tokens that are the same to the word 'x' in writing position. For instance, the value of 0.95 % for the field {n,m-1} represents the chance that two consecutive words are identical, whereas the value of 0.99 % in {n-1,m} represents the chance that a word is identical to the word in the previous line in the same position.

The table shows that there is an increased chance that a word is identical to one of the previous words in the same line or to the words in a similar position in one of the previous lines. In other words, the chance that a word is identical to another word decreases with the distance between the two words. This happens although features typical for a certain position in a paragraph or line exist. For instance, the ⟨-glyph ("m") occurs in 62 % of the cases at the end of a line [Timm: 19]. Moreover, the words at the beginnings of the lines are, on average, longer whereas the words at the end of the lines are, on average shorter than the average word length for the whole VMS [Timm: 35].

---

[4] By using the EVA alphabet it is possible to write the VMS-word ccog as "chol". The EVA alphabet, created by René Zandbergen and Gabriel Landini, can be used to analyze the text and to name the VMS-words [see Zandbergen]. The letters do not give any information about the meaning of the corresponding VMS glyph.

[5] The transcription by Takeshi Takahashi [Takahashi] was used for the statistical evaluation, whereby all single letter glyph sequences were eliminated. In order to keep the algorithm simple the whole book was treated as a unit. The first line of a page was compared with the last line of the previous page and so on.

|     | m-6   | m-5   | m-4   | m-3   | m-2   | m-1   | m     | m+1   | m+2   | m+3   | m+4   | m+5   | m+6   |
|-----|-------|-------|-------|-------|-------|-------|-------|-------|-------|-------|-------|-------|-------|
| n-9 | 1.62% | 1.54% | 1.64% | 1.84% | 1.88% | 1.99% | 2.04% | 1.87% | 1.93% | 1.87% | 1.74% | 1.67% | 1.68% |
| n-8 | 1.48% | 1.66% | 1.76% | 1.55% | 1.63% | 1.89% | 2.01% | 1.84% | 1.89% | 1.75% | 1.67% | 1.67% | 1.80% |
| n-7 | 1.90% | 1.84% | 1.90% | 1.70% | 1.93% | 2.14% | 2.19% | 2.00% | 1.77% | 1.77% | 1.82% | 1.84% | 1.78% |
| n-6 | 1.89% | 1.84% | 1.81% | 1.66% | 1.96% | 2.02% | 2.02% | 1.96% | 1.88% | 1.84% | 1.83% | 1.74% | 1.73% |
| n-5 | 1.50% | 1.53% | 1.83% | 1.76% | 2.02% | 1.94% | 2.02% | 2.07% | 2.06% | 1.79% | 2.02% | 1.79% | 1.57% |
| n-4 | 1.63% | 1.87% | 1.90% | 1.68% | 2.00% | 2.15% | 2.01% | 2.14% | 2.03% | 1.83% | 1.68% | 1.70% | 1.69% |
| n-3 | 1.76% | 1.87% | 1.81% | 1.94% | 2.04% | 2.16% | 2.17% | 2.15% | 2.15% | 2.11% | 1.97% | 1.90% | 1.65% |
| n-2 | 1.93% | 1.78% | 2.02% | 2.06% | 2.13% | 2.30% | 2.29% | 2.12% | 2.11% | 1.90% | 1.85% | 1.74% | 1.93% |
| n-1 | 1.65% | 1.97% | 2.06% | 2.18% | 2.13% | 2.46% | 2.57% | 2.40% | 2.41% | 2.25% | 2.30% | 2.20% | 2.30% |
| n   |       | 2.11% | 2.20% | 2.32% | 2.33% | 2.70% | 2.82% | 'x'   |       |       |       |       |       |

Table 3: The VMS – proportion of similar words with an edit distance of one

Astonishingly, the same correlation can be found for similarly spelled glyph sequences. Table 3 gives the probabilities that two words with an edit distance of one occur near to each other. As in table 2 the value of 2.79 % in {n,m-1} stands for the chance that two consecutive glyph sequences differ in only one glyph.[6] Table 4 demonstrates a similar context dependency for words with an edit distance of two.

|     | m-6   | m-5   | m-4   | m-3   | m-2   | m-1   | m     | m+1   | m+2   | m+3   | m+4   | m+5   | m+6   |
|-----|-------|-------|-------|-------|-------|-------|-------|-------|-------|-------|-------|-------|-------|
| n-9 | 3.45% | 3.68% | 3.68% | 3.85% | 3.85% | 4.06% | 4.17% | 3.90% | 4.06% | 3.78% | 3.75% | 3.60% | 3.83% |
| n-8 | 3.24% | 3.72% | 3.58% | 3.77% | 3.83% | 4.17% | 4.11% | 4.01% | 3.86% | 3.66% | 3.69% | 3.52% | 3.53% |
| n-7 | 3.39% | 3.75% | 3.78% | 3.92% | 3.90% | 4.05% | 4.22% | 4.09% | 3.95% | 3.81% | 3.52% | 3.72% | 3.78% |
| n-6 | 3.77% | 3.72% | 3.80% | 3.91% | 4.02% | 4.06% | 4.32% | 4.15% | 3.97% | 3.89% | 3.80% | 4.13% | 3.85% |
| n-5 | 3.48% | 3.63% | 3.71% | 3.54% | 4.02% | 4.23% | 4.21% | 4.33% | 4.10% | 3.67% | 3.75% | 3.82% | 3.54% |
| n-4 | 3.41% | 3.75% | 3.93% | 3.85% | 4.06% | 4.22% | 4.29% | 4.23% | 3.94% | 4.13% | 3.79% | 3.79% | 3.84% |
| n-3 | 3.83% | 4.04% | 3.89% | 3.89% | 4.13% | 4.32% | 4.51% | 4.28% | 4.20% | 4.02% | 3.98% | 3.95% | 3.73% |
| n-2 | 3.94% | 3.88% | 3.82% | 4.20% | 4.25% | 4.41% | 4.56% | 4.48% | 4.19% | 4.29% | 3.85% | 4.00% | 3.70% |
| n-1 | 3.81% | 3.87% | 4.04% | 3.87% | 4.27% | 4.55% | 4.65% | 4.55% | 4.58% | 4.54% | 4.73% | 4.22% | 4.11% |
| n   |       | 3.84% | 4.32% | 4.36% | 4.74% | 5.08% | 4.87% | 'x'   |       |       |       |       |       |

Table 4: The VMS – proportion of similar words with an edit distance of two

Thus, all three tables demonstrate that there is an increased chance for the occurrence of identical or similar words near

---

[6] To consider the peculiarities of the VMS script, one change is defined as a glyph deleted, added or replaced by a similar glyph. Glyphs handled as similar are: o/a, o/9, 8/ɣ/ß/2, ɔ/ʔ/2 and ⟨/⟨/⟨/⟨. If a glyph is replaced by a non-similar glyph, this is counted as two changes - for deleting one glyph and adding another glyph.

to each other.[7] In some way the occurrence of a word increases the probability that the same or similar words occur in the same line or in one of the next lines in a similar position. For the VMS it is not only possible to predict how often a particular word exists but also that a given word can be found near to similar ones. For instance, the bigram ᴏᴎ ("on") is far less frequent (5 : 126) than the bigram ᴀᴎ ("an"). For this reason, it is not surprising that a word ᴄʜᴏɴ ("chon") occurs only once, whereas the word ᴄʜᴀɴ ("chan") occurs eleven times. What is surprising is that the word ᴄʜᴏɴ occurs on page <f37v> very close to words like ǫᴏᴋᴄʜᴏɴ ("qokchon") and ʏᴋᴄʜᴏɴ ("ykchon"):

| <f37v.P.3> | qokchon … chon | ("qokchon … chon") |
| <f37v.P.5> | ykchon | ("ykchon") |

This means that three out of five words containing the bigram ᴏɴ ("on") occurred within a space of only two lines. Moreover, these words also occur near to other similar words like ᴄʜᴏʀ ("chor"), ʏᴛᴄʜᴏʀ ("ytchor") and ǫᴏᴛᴄʜᴏʀ ("qotchor").[8]

|   | m-6 | m-5 | m-4 | m-3 | m-2 | m-1 | m | m+1 | m+2 | m+3 | m+4 | m+5 | m+6 |
|---|---|---|---|---|---|---|---|---|---|---|---|---|---|
| n-9 | 0.23% | 0.24% | 0.26% | 0.23% | 0.28% | 0.28% | 0.72% | 0.26% | 0.19% | 0.26% | 0.23% | 0.22% | 0.27% |
| n-8 | 0.20% | 0.27% | 0.27% | 0.19% | 0.24% | 0.27% | 0.74% | 0.26% | 0.23% | 0.25% | 0.23% | 0.21% | 0.26% |
| n-7 | 0.27% | 0.23% | 0.23% | 0.22% | 0.26% | 0.28% | 0.67% | 0.28% | 0.28% | 0.26% | 0.23% | 0.24% | 0.28% |
| n-6 | 0.24% | 0.26% | 0.25% | 0.26% | 0.24% | 0.24% | 0.90% | 0.32% | 0.28% | 0.20% | 0.25% | 0.31% | 0.30% |
| n-5 | 0.26% | 0.21% | 0.22% | 0.28% | 0.30% | 0.30% | 0.84% | 0.31% | 0.25% | 0.28% | 0.22% | 0.25% | 0.26% |
| n-4 | 0.30% | 0.24% | 0.23% | 0.26% | 0.29% | 0.27% | 1.03% | 0.33% | 0.28% | 0.22% | 0.24% | 0.29% | 0.31% |
| n-3 | 0.21% | 0.26% | 0.30% | 0.25% | 0.32% | 0.36% | 0.96% | 0.30% | 0.27% | 0.30% | 0.31% | 0.25% | 0.31% |
| n-2 | 0.28% | 0.28% | 0.28% | 0.27% | 0.28% | 0.37% | 1.30% | 0.34% | 0.33% | 0.36% | 0.33% | 0.33% | 0.32% |
| n-1 | 0.29% | 0.32% | 0.32% | 0.41% | 0.37% | 0.43% | 1.52% | 0.52% | 0.44% | 0.38% | 0.35% | 0.37% | 0.45% |
| n | 0.61% | 0.68% | 0.66% | 0.62% | 0.26% | 0.01% | 'x' |  |  |  |  |  |  |

**Table 5:** Arabic Quran – proportion of identical words appearing near to the word in writing position 'x'

The Arabic Quran (in Buckwalter transcription) was used to compare the VMS text with natural language.[9] Table 5 demonstrates the proportions of identical words appearing near to the word in imaginary writing position 'x'. As expected, some features point to context dependency, but the details show different results. The value of 0.01 % for consecutive words {n,m-1} demonstrates the fact that natural languages usually avoid using the same word twice in a row. However, the Quran is similar to the VMS in that the probabilities of a word being repeated are increased for the same line and for the next lines. This is the case because in natural languages and in the VMS the words depend on their context. The difference is that in the Arabic text the words prefer a certain position within a line, whereas in the VMS the same or similar words only occur near to each other.

---

[7] The same pattern can be demonstrated for parts of the VMS. See addendum for tables for pages using Currier A (p. 13), Currier B (p. 14), pages with "Herbal" illustrations in Currier A (p. 15), pages with "Herbal" illustrations in Currier B (p. 16), the "Biological" section (p. 17) and the "Stars" section (p. 18). If line-initial and line-terminal words where removed the pattern also remains unchanged (see p. 19).

[8] The word ᴄʜᴏʀ ("chor") occurs in line <f37v.P.2>, the word ʏᴛᴄʜᴏʀ ("ytchor") in line <f37v.P.6> and the word ǫᴏᴛᴄʜᴏʀ ("qotchor") in line <f37v.P.7>. The type ᴄʜᴏʀ occurs 219 times, and ʏᴛᴄʜᴏʀ and ǫᴏᴛᴄʜᴏʀ occur thirteen and fourteen times, respectively, within the VMS.

[9] Similar results would be obtained by comparing the VMS with poetry in other languages. See addendum p. 10 for English poetry and p. 11 for Latin poetry.

A different result would be obtained for a continuous text. In such a text, similar proportions of identical words are obtained for all positions.[10] This happens since for such a text a word can occur in any position within a line.

Table 6 gives the percentages for the occurrences of words in the Arabic text where one letter is different compared to the word 'x' in writing position. The preference for a certain position within a line for similar words is weaker than for the same words. The number of similar words is approximately on the same level as in table 5, whereas within the VMS there was a far higher co-occurrence of similar words than of the same words.

|     | m-6   | m-5   | m-4   | m-3   | m-2   | m-1   | m     | m+1   | m+2   | m+3   | m+4   | m+5   | m+6   |
| --- | ----- | ----- | ----- | ----- | ----- | ----- | ----- | ----- | ----- | ----- | ----- | ----- | ----- |
| n-9 | 0.30% | 0.31% | 0.30% | 0.27% | 0.29% | 0.30% | 0.36% | 0.33% | 0.28% | 0.26% | 0.32% | 0.32% | 0.26% |
| n-8 | 0.30% | 0.28% | 0.29% | 0.31% | 0.27% | 0.31% | 0.36% | 0.29% | 0.30% | 0.33% | 0.30% | 0.33% | 0.25% |
| n-7 | 0.26% | 0.22% | 0.32% | 0.31% | 0.25% | 0.29% | 0.38% | 0.32% | 0.29% | 0.33% | 0.28% | 0.34% | 0.30% |
| n-6 | 0.23% | 0.30% | 0.30% | 0.26% | 0.27% | 0.29% | 0.40% | 0.25% | 0.29% | 0.30% | 0.32% | 0.29% | 0.30% |
| n-5 | 0.35% | 0.28% | 0.34% | 0.27% | 0.35% | 0.26% | 0.40% | 0.28% | 0.28% | 0.28% | 0.26% | 0.33% | 0.30% |
| n-4 | 0.28% | 0.25% | 0.27% | 0.30% | 0.26% | 0.35% | 0.37% | 0.30% | 0.27% | 0.29% | 0.29% | 0.35% | 0.32% |
| n-3 | 0.32% | 0.34% | 0.31% | 0.30% | 0.30% | 0.36% | 0.39% | 0.32% | 0.27% | 0.33% | 0.29% | 0.29% | 0.31% |
| n-2 | 0.27% | 0.30% | 0.31% | 0.33% | 0.26% | 0.33% | 0.42% | 0.29% | 0.34% | 0.35% | 0.32% | 0.30% | 0.30% |
| n-1 | 0.33% | 0.28% | 0.30% | 0.32% | 0.31% | 0.35% | 0.46% | 0.36% | 0.34% | 0.31% | 0.32% | 0.32% | 0.36% |
| n   | 0.40% | 0.41% | 0.45% | 0.39% | 0.33% | 0.09% | 'x'   |       |       |       |       |       |       |

Table 6: Arabic Quran – proportion of similar words with an edit distance of one

## Discussion

Two important differences were found between the VMS text and the Arabic text. First, the same and similar words are used near to each other in the VMS text, whereas in the Arabic text the words are used in a certain position; moreover, it is avoided to use the same or similar words in a row. Second, the level of context dependency for the VMS text is more compre-hensive since the same, and also similar, words were used near to each other.

---

[10] See addendum p. 12 for the proportions for a continuous text in German.

Montemurro and Zanette argue with respect to the context dependency: "Words that are related by their semantic contents tend to co-occur along the text." [Montemurro]. They come to the conclusion that this observation "suggests the presence of a genuine linguistic structure" for the VMS. Unfortunately, the level of context dependency is on a higher level than expected for a linguistic system. Therefore, context dependency alone is not enough to allow the conclusion that the patterns found must be the result of a genuine linguistic structure.

The words in the VMS build a network of words similar to each other.[11] Therefore it is no surprise that a larger number of similar words exist for each word in the VMS. However, they occur near to each other and they occur with comparable frequencies. How was it possible to construct a language with "generated" words and to write a text containing over 37,000 words with determinable word frequencies? Was the scribe counting the words he was writing? A better explanation is that since similar words do co-occur throughout the text, the spelling variations for a frequently used word also occur more often. In other words, for the VMS, the observed word frequencies are a result of the fact that similar words do co-occur throughout the text.

In natural languages a word normally (cf. poems) is used because of its meaning and not because it is similar to a previously written one. The result that the words are arranged such that they co-occur with similar ones is therefore not compatible with a linguistic system. An English text with similar features would mainly consist of words similar to the words "the", "and" and "to". Additionally, a word "the" would co-occur with words like "khe", "phe", "fhe", "tha", "tho", "thy", "thee" and "theee".

The features described above point to a text generation algorithm. A hypothesis describing such an algorithm was published in 2014 [see Timm]. According to this hypothesis, the whole manuscript consists of permutations of the three most commonly used glyph sequences ꝼaɩɩƆ ("daiin"), oɤ ("ol") and ccꝼꝿ ("chedy").[12]

One outstanding feature of the VMS is that in 86% of cases the first word of a paragraph starts with an additional gallow glyph. If the scribe preferred glyph sequences previously written as a source for generating new text it could be expected that these additional gallow glyphs would have an effect on the text. It is indeed possible to describe such effects.

---

[11] This is at least the case for words used four or more times [see Timm: 66-82].

[12] Connections between these three series exist. It is possible to demonstrate multiple paths between ꝼaɩɩƆ ("daiin") and oɤ ("ol") or between ꝼaɩɩƆ ("daiin") and ccꝼꝿ ("chedy"). An example for a path between ꝼaɩɩƆ and oɤ is ꝼaɩɩƆ – ꝼaɩƆ – ꝼaƆ – ꝼaᒿ – aᒿ – aɤ – oɤ. An example for a path between ꝼaɩɩƆ and ccꝼꝿ is ꝼaɩɩƆ – ccꝼaɩɩƆ – ccꝼaɩɩƆ – ccꝼaɩƆ – ccꝼaƆ – ccꝼa – ccꝼꝿ.

First, the paragraph-initial lines contain more words using gallow glyphs. After adding the paragraph-initial gallow glyph one can imagine that if the scribe had an idea of adding a gallow glyph in his mind, there would have been an increased chance that he would add further gallow glyphs [see Timm: p. 30].

Second, the line-initial words are, on average, longer because of the repercussion of the paragraph-initial gallow glyph. The first and the last word in each line are easy to spot. The most obvious way to proceed is, therefore, to pick them as a source for the generation of a word at the beginning or at the end of a line. Therefore, the first word in a line commonly starts with a prefix ꝯ, o, ß or ꝛ. For the second word it is also possible to select the first word as a source. An obvious change in such a case would be to remove the additional prefix. And indeed, the second word in a line occurs twice as frequent as a subgroup of the preceding word (2.6%) than this is the case for the words in other positions (1.3%) [Timm: 20]. In other words, the second glyph sequence in a line is shorter on average, since the first glyph sequence is longer on average.

To change glyph sequences which have already been written requires no additional tools. For someone in the early 15$^{th}$ century it would therefore have been possible to use such a method to generate the text of the Voynich Manuscript while writing.

**Addendum**

```
I. English poetry – The Towneley plays: The Creation
```

| | m-5 | m-4 | m-3 | m-2 | m-1 | m | m+1 | m+2 | m+3 | m+4 | m+5 |
|---|---|---|---|---|---|---|---|---|---|---|---|
| n-9 | 0.42% | 0.43% | 0.52% | 0.64% | 0.66% | 1.55% | 0.61% | 0.65% | 0.51% | 0.49% | 0.52% |
| n-8 | 0.36% | 0.44% | 0.54% | 0.70% | 0.71% | 1.80% | 0.69% | 0.64% | 0.57% | 0.48% | 0.47% |
| n-7 | 0.30% | 0.58% | 0.57% | 0.60% | 0.67% | 1.78% | 0.73% | 0.66% | 0.51% | 0.56% | 0.43% |
| n-6 | 0.41% | 0.52% | 0.57% | 0.62% | 0.62% | 1.82% | 0.72% | 0.65% | 0.57% | 0.50% | 0.42% |
| n-5 | 0.53% | 0.50% | 0.59% | 0.62% | 0.74% | 1.75% | 0.68% | 0.62% | 0.57% | 0.56% | 0.52% |
| n-4 | 0.58% | 0.55% | 0.51% | 0.70% | 0.83% | 1.94% | 0.78% | 0.71% | 0.60% | 0.55% | 0.45% |
| n-3 | 0.47% | 0.49% | 0.61% | 0.71% | 0.74% | 1.73% | 0.90% | 0.74% | 0.56% | 0.56% | 0.51% |
| n-2 | 0.51% | 0.63% | 0.62% | 0.77% | 0.89% | 1.81% | 0.88% | 0.80% | 0.65% | 0.53% | 0.46% |
| n-1 | 0.54% | 0.70% | 0.67% | 0.79% | 0.97% | 1.56% | 0.91% | 0.84% | 0.73% | 0.70% | 0.54% |
| n | 0.65% | 0.90% | 0.60% | 0.41% | 0.04% | 'x' | | | | | |

**Table 7: English poetry[13] – proportion of identical words appearing near to the word in writing position 'x'**

| | m-5 | m-4 | m-3 | m-2 | m-1 | m | m+1 | m+2 | m+3 | m+4 | m+5 |
|---|---|---|---|---|---|---|---|---|---|---|---|
| n-9 | 1.10% | 1.18% | 1.34% | 1.34% | 1.25% | 1.33% | 1.29% | 1.32% | 1.25% | 1.30% | 1.31% |
| n-8 | 1.25% | 1.38% | 1.30% | 1.46% | 1.34% | 1.30% | 1.41% | 1.37% | 1.27% | 1.47% | 1.30% |
| n-7 | 1.09% | 1.30% | 1.29% | 1.28% | 1.28% | 1.36% | 1.43% | 1.37% | 1.22% | 1.22% | 1.32% |
| n-6 | 1.24% | 1.40% | 1.33% | 1.43% | 1.35% | 1.44% | 1.42% | 1.45% | 1.25% | 1.21% | 1.36% |
| n-5 | 1.26% | 1.44% | 1.35% | 1.35% | 1.31% | 1.41% | 1.39% | 1.36% | 1.33% | 1.34% | 1.29% |
| n-4 | 1.13% | 1.33% | 1.40% | 1.60% | 1.50% | 1.52% | 1.48% | 1.39% | 1.32% | 1.42% | 1.38% |
| n-3 | 1.35% | 1.39% | 1.32% | 1.46% | 1.55% | 1.59% | 1.64% | 1.49% | 1.34% | 1.23% | 1.31% |
| n-2 | 1.31% | 1.36% | 1.40% | 1.64% | 1.89% | 2.07% | 1.89% | 1.59% | 1.38% | 1.31% | 1.14% |
| n-1 | 1.24% | 1.18% | 1.41% | 1.50% | 1.77% | 1.89% | 1.86% | 1.51% | 1.31% | 1.21% | 1.24% |
| n | 1.43% | 1.38% | 1.36% | 1.40% | 1.22% | 'x' | | | | | |

**Table 8: English poetry[13] – proportion of similar words with an edit distance of one**

---

[13] `The Towneley Plays: The Creation [Anonymous]`

```
II. Latin poetry
```

|     | m-5   | m-4   | m-3   | m-2   | m-1   | m     | m+1   | m+2   | m+3   | m+4   | m+5   |
|-----|-------|-------|-------|-------|-------|-------|-------|-------|-------|-------|-------|
| n-9 | 0.10% | 0.12% | 0.17% | 0.23% | 0.24% | 0.27% | 0.24% | 0.21% | 0.17% | 0.20% | 0.06% |
| n-8 | 0.13% | 0.18% | 0.18% | 0.19% | 0.26% | 0.28% | 0.24% | 0.22% | 0.20% | 0.12% | 0.10% |
| n-7 | 0.13% | 0.18% | 0.20% | 0.20% | 0.21% | 0.32% | 0.24% | 0.23% | 0.23% | 0.16% | 0.14% |
| n-6 | 0.11% | 0.17% | 0.12% | 0.22% | 0.28% | 0.24% | 0.21% | 0.25% | 0.16% | 0.16% | 0.15% |
| n-5 | 0.07% | 0.14% | 0.16% | 0.24% | 0.28% | 0.28% | 0.26% | 0.24% | 0.19% | 0.13% | 0.12% |
| n-4 | 0.12% | 0.15% | 0.19% | 0.22% | 0.24% | 0.27% | 0.26% | 0.25% | 0.20% | 0.17% | 0.08% |
| n-3 | 0.12% | 0.16% | 0.18% | 0.24% | 0.28% | 0.26% | 0.22% | 0.23% | 0.23% | 0.12% | 0.08% |
| n-2 | 0.10% | 0.14% | 0.16% | 0.20% | 0.27% | 0.27% | 0.24% | 0.21% | 0.14% | 0.16% | 0.14% |
| n-1 | 0.08% | 0.12% | 0.15% | 0.24% | 0.25% | 0.29% | 0.24% | 0.20% | 0.15% | 0.17% | 0.20% |
| n   | 0.28% | 0.44% | 0.47% | 0.31% | 0.03% | 'x'   |       |       |       |       |       |

**Table 9: Latin poetry[14] – proportion of identical words appearing near to the word in writing position 'x'**

|     | m-5   | m-4   | m-3   | m-2   | m-1   | m     | m+1   | m+2   | m+3   | m+4   | m+5   |
|-----|-------|-------|-------|-------|-------|-------|-------|-------|-------|-------|-------|
| n-9 | 0.14% | 0.23% | 0.23% | 0.21% | 0.22% | 0.29% | 0.22% | 0.23% | 0.17% | 0.20% | 0.20% |
| n-8 | 0.11% | 0.17% | 0.16% | 0.18% | 0.25% | 0.29% | 0.22% | 0.25% | 0.19% | 0.22% | 0.20% |
| n-7 | 0.20% | 0.19% | 0.22% | 0.22% | 0.21% | 0.26% | 0.25% | 0.20% | 0.28% | 0.20% | 0.16% |
| n-6 | 0.23% | 0.13% | 0.20% | 0.23% | 0.21% | 0.27% | 0.21% | 0.20% | 0.21% | 0.15% | 0.15% |
| n-5 | 0.10% | 0.20% | 0.21% | 0.23% | 0.24% | 0.27% | 0.22% | 0.22% | 0.21% | 0.22% | 0.15% |
| n-4 | 0.19% | 0.17% | 0.21% | 0.23% | 0.25% | 0.30% | 0.26% | 0.24% | 0.22% | 0.21% | 0.22% |
| n-3 | 0.15% | 0.26% | 0.18% | 0.26% | 0.25% | 0.26% | 0.21% | 0.23% | 0.25% | 0.22% | 0.26% |
| n-2 | 0.19% | 0.22% | 0.19% | 0.22% | 0.23% | 0.29% | 0.23% | 0.16% | 0.22% | 0.24% | 0.19% |
| n-1 | 0.15% | 0.23% | 0.20% | 0.18% | 0.23% | 0.25% | 0.19% | 0.21% | 0.18% | 0.12% | 0.17% |
| n   | 0.21% | 0.25% | 0.18% | 0.17% | 0.09% | 'x'   |       |       |       |       |       |

**Table 10: Latin poetry[14] – proportion of similar words with an edit distance of one**

---

[14] The Aeneid by Virgil 80–19 BC [Virgil]

## III. German continuous text

|     | m-5   | m-4   | m-3   | m-2   | m-1   | m     | m+1   | m+2   | m+3   | m+4   | m+5   |
|-----|-------|-------|-------|-------|-------|-------|-------|-------|-------|-------|-------|
| n-9 | 0.56% | 0.57% | 0.56% | 0.58% | 0.56% | 0.56% | 0.56% | 0.58% | 0.58% | 0.55% | 0.57% |
| n-8 | 0.59% | 0.58% | 0.58% | 0.53% | 0.57% | 0.59% | 0.57% | 0.57% | 0.57% | 0.57% | 0.58% |
| n-7 | 0.58% | 0.55% | 0.58% | 0.60% | 0.58% | 0.58% | 0.60% | 0.56% | 0.57% | 0.57% | 0.58% |
| n-6 | 0.58% | 0.58% | 0.60% | 0.59% | 0.58% | 0.59% | 0.57% | 0.61% | 0.59% | 0.60% | 0.61% |
| n-5 | 0.59% | 0.59% | 0.59% | 0.62% | 0.60% | 0.59% | 0.61% | 0.59% | 0.60% | 0.61% | 0.61% |
| n-4 | 0.63% | 0.59% | 0.60% | 0.62% | 0.60% | 0.64% | 0.61% | 0.59% | 0.58% | 0.61% | 0.61% |
| n-3 | 0.61% | 0.62% | 0.63% | 0.62% | 0.63% | 0.65% | 0.64% | 0.62% | 0.61% | 0.63% | 0.66% |
| n-2 | 0.62% | 0.64% | 0.67% | 0.65% | 0.69% | 0.65% | 0.67% | 0.66% | 0.66% | 0.67% | 0.67% |
| n-1 | 0.67% | 0.67% | 0.66% | 0.67% | 0.70% | 0.69% | 0.69% | 0.69% | 0.68% | 0.66% | 0.66% |
| n   | 0.71% | 0.63% | 0.54% | 0.31% | 0.01% | 'x'   |       |       |       |       |       |

**Table 11: German continuous text[15] – proportion of identical words**

|     | m-5   | m-4   | m-3   | m-2   | m-1   | m     | m+1   | m+2   | m+3   | m+4   | m+5   |
|-----|-------|-------|-------|-------|-------|-------|-------|-------|-------|-------|-------|
| n-9 | 0.75% | 0.73% | 0.74% | 0.77% | 0.76% | 0.75% | 0.74% | 0.75% | 0.73% | 0.80% | 0.75% |
| n-8 | 0.78% | 0.76% | 0.75% | 0.75% | 0.78% | 0.76% | 0.75% | 0.78% | 0.74% | 0.76% | 0.77% |
| n-7 | 0.74% | 0.76% | 0.75% | 0.76% | 0.75% | 0.76% | 0.74% | 0.76% | 0.77% | 0.77% | 0.77% |
| n-6 | 0.75% | 0.77% | 0.77% | 0.77% | 0.76% | 0.77% | 0.78% | 0.76% | 0.78% | 0.73% | 0.76% |
| n-5 | 0.75% | 0.76% | 0.75% | 0.77% | 0.74% | 0.73% | 0.73% | 0.76% | 0.77% | 0.74% | 0.75% |
| n-4 | 0.80% | 0.74% | 0.79% | 0.74% | 0.78% | 0.77% | 0.75% | 0.76% | 0.74% | 0.77% | 0.79% |
| n-3 | 0.76% | 0.77% | 0.76% | 0.77% | 0.79% | 0.73% | 0.75% | 0.77% | 0.79% | 0.81% | 0.78% |
| n-2 | 0.78% | 0.75% | 0.78% | 0.77% | 0.78% | 0.76% | 0.77% | 0.77% | 0.77% | 0.78% | 0.79% |
| n-1 | 0.77% | 0.81% | 0.77% | 0.82% | 0.80% | 0.79% | 0.75% | 0.78% | 0.76% | 0.76% | 0.75% |
| n   | 0.75% | 0.71% | 0.73% | 0.70% | 0.33% | 'x'   |       |       |       |       |       |

**Table 12: German continuous text[15] – proportion of similar words with an edit distance of one**

---

[15] Illustrirtes Thierleben. Bd. I by Alfred Edmund Brehm [Brehm]

## IV. VMS pages in Currier A

|     | m-6   | m-5   | m-4   | m-3   | m-2   | m-1   | m     | m+1   | m+2   | m+3   | m+4   | m+5   | m+6   |
|-----|-------|-------|-------|-------|-------|-------|-------|-------|-------|-------|-------|-------|-------|
| n-9 | 0.55% | 0.39% | 0.47% | 0.54% | 0.48% | 0.78% | 0.87% | 0.78% | 0.72% | 0.59% | 0.54% | 0.51% | 0.67% |
| n-8 | 0.75% | 0.67% | 0.71% | 0.64% | 0.61% | 0.57% | 0.63% | 0.78% | 0.75% | 0.66% | 0.60% | 0.93% | 0.44% |
| n-7 | 0.34% | 0.52% | 0.55% | 0.76% | 0.57% | 0.67% | 0.85% | 0.72% | 0.66% | 0.97% | 0.67% | 0.59% | 0.63% |
| n-6 | 0.39% | 0.48% | 0.72% | 0.71% | 0.75% | 0.63% | 0.87% | 0.80% | 0.87% | 0.59% | 0.54% | 0.62% | 0.68% |
| n-5 | 0.48% | 0.72% | 0.59% | 0.97% | 0.63% | 0.79% | 0.76% | 0.70% | 0.82% | 0.73% | 0.95% | 0.69% | 0.39% |
| n-4 | 0.66% | 0.67% | 0.61% | 0.77% | 0.75% | 0.76% | 0.93% | 0.65% | 0.74% | 0.74% | 0.63% | 0.64% | 0.62% |
| n-3 | 0.41% | 0.78% | 0.66% | 0.62% | 0.63% | 0.99% | 1.03% | 0.71% | 0.59% | 0.54% | 0.90% | 0.58% | 0.55% |
| n-2 | 0.52% | 0.84% | 0.68% | 0.91% | 0.69% | 0.83% | 1.04% | 0.80% | 0.94% | 0.81% | 0.93% | 0.86% | 0.73% |
| n-1 | 0.66% | 0.80% | 0.67% | 0.65% | 0.99% | 0.84% | 0.99% | 1.17% | 1.06% | 0.71% | 1.12% | 0.70% | 1.06% |
| n   | 0.86% | 0.62% | 1.03% | 1.02% | 1.01% | 1.13% | 'x'   |       |       |       |       |       |       |

**Table 13: VMS Currier A[16] – proportion of identical words appearing near to the word in writing position 'x'**

|     | m-6   | m-5   | m-4   | m-3   | m-2   | m-1   | m     | m+1   | m+2   | m+3   | m+4   | m+5   | m+6   |
|-----|-------|-------|-------|-------|-------|-------|-------|-------|-------|-------|-------|-------|-------|
| n-9 | 1.15% | 1.48% | 1.58% | 1.71% | 1.49% | 1.91% | 2.03% | 1.78% | 1.65% | 1.70% | 1.76% | 1.67% | 1.78% |
| n-8 | 1.10% | 1.44% | 1.79% | 1.30% | 1.36% | 1.65% | 1.97% | 1.57% | 1.96% | 1.87% | 1.77% | 1.38% | 1.41% |
| n-7 | 1.72% | 1.46% | 1.35% | 1.74% | 1.60% | 2.00% | 1.88% | 1.97% | 1.64% | 1.73% | 1.43% | 1.70% | 1.42% |
| n-6 | 1.36% | 1.28% | 1.76% | 1.52% | 1.62% | 1.97% | 2.11% | 2.18% | 1.50% | 1.62% | 1.81% | 2.27% | 1.32% |
| n-5 | 1.30% | 1.23% | 1.62% | 1.53% | 1.95% | 1.73% | 1.93% | 1.85% | 2.06% | 1.56% | 1.88% | 1.68% | 1.47% |
| n-4 | 1.08% | 1.75% | 1.62% | 1.59% | 1.87% | 1.99% | 1.98% | 1.88% | 1.84% | 1.59% | 1.57% | 1.49% | 1.39% |
| n-3 | 2.19% | 1.40% | 1.74% | 1.88% | 1.72% | 1.88% | 2.14% | 1.99% | 2.28% | 1.65% | 1.84% | 1.91% | 1.10% |
| n-2 | 1.79% | 1.63% | 1.72% | 1.82% | 2.00% | 2.17% | 2.12% | 2.05% | 1.94% | 1.83% | 1.90% | 2.09% | 1.68% |
| n-1 | 1.19% | 2.18% | 1.81% | 2.02% | 1.84% | 2.44% | 2.41% | 2.59% | 2.44% | 2.31% | 2.38% | 2.07% | 2.40% |
| n   | 2.16% | 2.10% | 2.07% | 2.30% | 2.82% | 3.24% | 'x'   |       |       |       |       |       |       |

**Table 14: VMS Currier A[16] – proportion of similar words with an edit distance of one**

|     | m-6   | m-5   | m-4   | m-3   | m-2   | m-1   | m     | m+1   | m+2   | m+3   | m+4   | m+5   | m+6   |
|-----|-------|-------|-------|-------|-------|-------|-------|-------|-------|-------|-------|-------|-------|
| n-9 | 3.46% | 3.56% | 3.73% | 4.14% | 3.87% | 3.96% | 4.27% | 3.95% | 4.40% | 4.25% | 3.75% | 3.75% | 3.71% |
| n-8 | 2.94% | 3.48% | 3.31% | 3.24% | 4.05% | 4.24% | 4.10% | 4.42% | 3.96% | 3.94% | 3.93% | 3.14% | 3.64% |
| n-7 | 3.30% | 3.62% | 3.80% | 4.12% | 3.68% | 4.49% | 4.51% | 4.57% | 4.29% | 3.82% | 3.74% | 3.77% | 4.15% |
| n-6 | 3.41% | 3.59% | 3.78% | 3.80% | 4.64% | 4.48% | 4.64% | 4.22% | 4.00% | 4.04% | 3.89% | 4.05% | 4.00% |
| n-5 | 3.57% | 4.04% | 3.65% | 3.83% | 4.17% | 4.67% | 4.61% | 4.38% | 4.25% | 4.00% | 3.73% | 4.05% | 3.52% |
| n-4 | 3.62% | 4.00% | 4.12% | 3.89% | 4.29% | 4.30% | 4.86% | 4.47% | 4.23% | 4.32% | 3.65% | 4.30% | 4.37% |
| n-3 | 3.92% | 4.42% | 4.04% | 3.71% | 4.14% | 4.40% | 4.88% | 4.37% | 4.41% | 4.23% | 3.98% | 3.94% | 3.71% |
| n-2 | 3.36% | 4.06% | 3.54% | 4.11% | 4.21% | 4.56% | 4.86% | 4.43% | 4.73% | 4.36% | 4.50% | 3.97% | 4.05% |
| n-1 | 3.81% | 4.13% | 4.06% | 3.75% | 4.63% | 4.99% | 5.10% | 5.18% | 4.97% | 4.66% | 4.83% | 4.77% | 4.63% |
| n   | 4.00% | 4.68% | 4.81% | 5.36% | 5.04% | 5.74% | 'x'   |       |       |       |       |       |       |

**Table 15: VMS Currier A[16] – proportion of similar words with an edit distance of two**

---

[16] Pages F1r–F25v; F27r–F30v; F32r–F32v; F35r–F38v; F42r–F42v; F44r–F45v; F47r–F47v; F49r–F49v; F51r–F54v; F56r–F56v; F58r–F58v; F87r–F93v; F96r–F102v2

# V. VMS pages in Currier B

|     | m-6   | m-5   | m-4   | m-3   | m-2   | m-1   | m     | m+1   | m+2   | m+3   | m+4   | m+5   | m+6   |
|-----|-------|-------|-------|-------|-------|-------|-------|-------|-------|-------|-------|-------|-------|
| n-9 | 0.68% | 0.56% | 0.56% | 0.65% | 0.65% | 0.80% | 0.79% | 0.90% | 0.74% | 0.65% | 0.68% | 0.69% | 0.51% |
| n-8 | 0.47% | 0.49% | 0.68% | 0.75% | 0.84% | 0.73% | 0.74% | 0.71% | 0.79% | 0.87% | 0.74% | 0.55% | 0.74% |
| n-7 | 0.48% | 0.42% | 0.71% | 0.72% | 0.73% | 0.70% | 0.72% | 0.78% | 0.81% | 0.65% | 0.78% | 0.60% | 0.57% |
| n-6 | 0.60% | 0.55% | 0.73% | 0.68% | 0.80% | 0.71% | 0.79% | 0.70% | 0.91% | 0.82% | 0.79% | 0.76% | 0.51% |
| n-5 | 0.54% | 0.52% | 0.62% | 0.78% | 0.68% | 0.86% | 0.94% | 0.76% | 0.74% | 0.73% | 0.60% | 0.64% | 0.62% |
| n-4 | 0.69% | 0.58% | 0.66% | 0.82% | 0.82% | 0.87% | 0.82% | 0.67% | 0.69% | 0.68% | 0.75% | 0.62% | 0.42% |
| n-3 | 0.62% | 0.64% | 0.63% | 0.66% | 0.73% | 0.95% | 0.80% | 0.84% | 0.85% | 0.77% | 0.57% | 0.79% | 0.74% |
| n-2 | 0.62% | 0.72% | 0.73% | 0.87% | 0.97% | 0.95% | 0.96% | 1.08% | 0.97% | 0.92% | 0.76% | 0.62% | 0.59% |
| n-1 | 0.82% | 0.71% | 0.72% | 0.83% | 0.94% | 0.95% | 1.05% | 0.95% | 1.00% | 0.83% | 0.88% | 0.84% | 0.73% |
| n   | 0.75% | 0.71% | 1.10% | 1.09% | 1.17% | 0.97% | 'x'   |       |       |       |       |       |       |

**Table 16: VMS Currier B[17] – proportion of identical words appearing near to the word in writing position 'x'**

|     | m-6   | m-5   | m-4   | m-3   | m-2   | m-1   | m     | m+1   | m+2   | m+3   | m+4   | m+5   | m+6   |
|-----|-------|-------|-------|-------|-------|-------|-------|-------|-------|-------|-------|-------|-------|
| n-9 | 1.68% | 1.52% | 1.70% | 1.92% | 2.00% | 2.00% | 2.16% | 1.97% | 2.06% | 1.94% | 1.83% | 1.70% | 1.72% |
| n-8 | 1.60% | 1.76% | 1.79% | 1.65% | 1.78% | 2.02% | 2.14% | 1.98% | 1.91% | 1.76% | 1.67% | 1.86% | 1.79% |
| n-7 | 1.93% | 1.95% | 2.08% | 1.70% | 2.11% | 2.22% | 2.36% | 2.09% | 1.82% | 1.87% | 1.98% | 2.01% | 1.91% |
| n-6 | 2.05% | 2.09% | 1.91% | 1.76% | 2.15% | 2.05% | 2.04% | 1.98% | 2.14% | 1.94% | 1.87% | 1.69% | 1.88% |
| n-5 | 1.61% | 1.65% | 1.93% | 1.86% | 2.08% | 2.08% | 2.18% | 2.27% | 2.17% | 1.95% | 2.13% | 1.89% | 1.66% |
| n-4 | 1.75% | 1.90% | 2.09% | 1.71% | 2.09% | 2.31% | 2.16% | 2.35% | 2.24% | 2.00% | 1.78% | 1.83% | 1.79% |
| n-3 | 1.66% | 1.94% | 1.86% | 1.96% | 2.19% | 2.33% | 2.26% | 2.29% | 2.18% | 2.30% | 2.07% | 1.96% | 1.79% |
| n-2 | 1.97% | 1.88% | 2.19% | 2.18% | 2.27% | 2.44% | 2.49% | 2.27% | 2.20% | 1.95% | 1.91% | 1.67% | 2.08% |
| n-1 | 1.83% | 1.93% | 2.25% | 2.29% | 2.31% | 2.55% | 2.65% | 2.40% | 2.48% | 2.27% | 2.43% | 2.33% | 2.41% |
| n   | 2.19% | 2.37% | 2.48% | 2.46% | 2.66% | 2.73% | 'x'   |       |       |       |       |       |       |

**Table 17: VMS Currier B[17] – proportion of similar words with an edit distance of one**

|     | m-6   | m-5   | m-4   | m-3   | m-2   | m-1   | m     | m+1   | m+2   | m+3   | m+4   | m+5   | m+6   |
|-----|-------|-------|-------|-------|-------|-------|-------|-------|-------|-------|-------|-------|-------|
| n-9 | 3.30% | 3.65% | 3.74% | 3.74% | 4.01% | 4.08% | 4.13% | 4.00% | 3.97% | 3.65% | 3.88% | 3.61% | 3.68% |
| n-8 | 3.25% | 3.83% | 3.67% | 3.99% | 3.79% | 4.26% | 4.17% | 3.92% | 3.86% | 3.69% | 3.66% | 3.74% | 3.65% |
| n-7 | 3.30% | 3.85% | 3.92% | 3.95% | 4.01% | 3.86% | 4.21% | 4.03% | 3.86% | 3.86% | 3.39% | 3.69% | 3.74% |
| n-6 | 3.75% | 3.77% | 3.98% | 3.99% | 3.93% | 3.88% | 4.29% | 4.22% | 4.06% | 3.90% | 3.84% | 4.02% | 3.83% |
| n-5 | 3.48% | 3.48% | 3.77% | 3.52% | 4.01% | 4.05% | 4.14% | 4.35% | 4.04% | 3.63% | 3.80% | 3.70% | 3.42% |
| n-4 | 3.45% | 3.70% | 3.87% | 3.77% | 4.07% | 4.30% | 4.14% | 4.20% | 3.91% | 4.19% | 4.01% | 3.74% | 3.79% |
| n-3 | 3.72% | 3.96% | 3.92% | 3.92% | 4.22% | 4.39% | 4.43% | 4.20% | 4.11% | 4.01% | 3.97% | 4.00% | 3.75% |
| n-2 | 3.94% | 3.79% | 3.94% | 4.25% | 4.34% | 4.36% | 4.55% | 4.50% | 4.07% | 4.37% | 3.66% | 3.95% | 3.70% |
| n-1 | 3.86% | 3.73% | 4.04% | 3.82% | 4.20% | 4.38% | 4.55% | 4.50% | 4.48% | 4.69% | 4.76% | 4.15% | 4.07% |
| n   | 3.89% | 4.31% | 4.42% | 4.45% | 5.05% | 4.56% | 'x'   |       |       |       |       |       |       |

**Table 18: VMS Currier B[17] – proportion of similar words with an edit distance of two**

---

[17] Pages F26r–F26v; F31r–F31v; F33r–F34v; F39r–F41v; F43r–F43v; F47r–F47v; F50r–F50v; F55r–F55v; F57r; F66r–F66v; F75r–F86v3; F94r–F95v1; F103r–F116r

# VI. VMS pages with "Herbal" illustrations in Currier A

|     | m-6   | m-5   | m-4   | m-3   | m-2   | m-1   | m     | m+1   | m+2   | m+3   | m+4   | m+5   | m+6   |
|-----|-------|-------|-------|-------|-------|-------|-------|-------|-------|-------|-------|-------|-------|
| n-9 | 0.80% | 0.47% | 0.48% | 0.53% | 0.53% | 0.87% | 0.97% | 0.82% | 0.87% | 0.61% | 0.55% | 0.60% | 0.69% |
| n-8 | 0.88% | 0.68% | 0.88% | 0.58% | 0.64% | 0.66% | 0.76% | 0.82% | 0.76% | 0.66% | 0.62% | 0.85% | 0.48% |
| n-7 | 0.56% | 0.73% | 0.69% | 0.88% | 0.64% | 0.71% | 0.89% | 0.81% | 0.73% | 1.08% | 0.58% | 0.66% | 0.77% |
| n-6 | 0.32% | 0.57% | 0.84% | 0.85% | 0.77% | 0.76% | 0.95% | 0.89% | 0.85% | 0.60% | 0.58% | 0.66% | 0.70% |
| n-5 | 0.63% | 0.87% | 0.73% | 1.06% | 0.57% | 0.83% | 0.85% | 0.62% | 0.86% | 0.71% | 1.00% | 0.82% | 0.43% |
| n-4 | 0.55% | 0.82% | 0.73% | 0.85% | 0.85% | 0.83% | 0.97% | 0.69% | 0.83% | 0.79% | 0.56% | 0.83% | 0.74% |
| n-3 | 0.47% | 0.97% | 0.69% | 0.63% | 0.76% | 1.11% | 1.04% | 0.87% | 0.60% | 0.58% | 0.94% | 0.59% | 0.68% |
| n-2 | 0.63% | 0.93% | 0.76% | 0.93% | 0.78% | 0.91% | 1.09% | 0.84% | 1.00% | 0.90% | 0.95% | 0.99% | 0.99% |
| n-1 | 0.55% | 0.86% | 0.78% | 0.64% | 1.12% | 1.04% | 1.07% | 1.26% | 1.17% | 0.71% | 1.12% | 0.74% | 1.39% |
| n   | 0.85% | 0.59% | 1.00% | 1.07% | 0.99% | 1.21% | 'x'   |       |       |       |       |       |       |

Table 19: VMS "Herbal" in Currier A[18] – proportion of identical words

|     | m-6   | m-5   | m-4   | m-3   | m-2   | m-1   | m     | m+1   | m+2   | m+3   | m+4   | m+5   | m+6   |
|-----|-------|-------|-------|-------|-------|-------|-------|-------|-------|-------|-------|-------|-------|
| n-9 | 1.45% | 1.72% | 1.54% | 1.66% | 1.38% | 2.06% | 2.08% | 1.88% | 1.58% | 1.63% | 1.81% | 1.58% | 1.92% |
| n-8 | 1.28% | 1.35% | 1.80% | 1.47% | 1.22% | 1.82% | 2.04% | 1.50% | 1.88% | 1.90% | 1.89% | 1.56% | 0.97% |
| n-7 | 1.91% | 1.66% | 1.39% | 1.96% | 1.57% | 2.09% | 2.05% | 2.02% | 1.61% | 1.66% | 1.44% | 1.94% | 1.19% |
| n-6 | 1.50% | 1.39% | 1.89% | 1.62% | 1.64% | 2.20% | 2.34% | 2.21% | 1.49% | 1.67% | 1.57% | 2.26% | 1.41% |
| n-5 | 1.33% | 1.18% | 1.78% | 1.66% | 2.10% | 1.81% | 1.92% | 1.89% | 2.10% | 1.61% | 1.86% | 1.68% | 1.38% |
| n-4 | 1.26% | 1.85% | 1.71% | 1.56% | 1.80% | 1.98% | 2.14% | 1.80% | 1.83% | 1.75% | 1.50% | 1.17% | 1.11% |
| n-3 | 1.81% | 1.38% | 1.85% | 1.88% | 1.82% | 1.77% | 2.30% | 1.97% | 2.28% | 1.75% | 1.88% | 1.42% | 1.05% |
| n-2 | 1.82% | 1.39% | 1.67% | 1.72% | 2.06% | 2.24% | 2.14% | 2.23% | 2.06% | 1.93% | 2.07% | 2.04% | 1.75% |
| n-1 | 1.40% | 2.07% | 1.39% | 1.81% | 1.83% | 2.31% | 2.45% | 2.76% | 2.50% | 2.17% | 2.42% | 2.03% | 2.62% |
| n   | 1.46% | 1.93% | 2.04% | 2.32% | 2.86% | 3.40% | 'x'   |       |       |       |       |       |       |

Table 20: VMS "Herbal" in Currier A[18] – proportion of similar words with an edit distance of one

|     | m-6   | m-5   | m-4   | m-3   | m-2   | m-1   | m     | m+1   | m+2   | m+3   | m+4   | m+5   | m+6   |
|-----|-------|-------|-------|-------|-------|-------|-------|-------|-------|-------|-------|-------|-------|
| n-9 | 3.21% | 3.60% | 3.45% | 3.82% | 4.18% | 4.06% | 4.55% | 3.97% | 4.55% | 4.30% | 3.96% | 3.95% | 3.84% |
| n-8 | 2.63% | 3.32% | 3.45% | 3.22% | 4.08% | 4.50% | 4.16% | 4.42% | 4.14% | 4.28% | 4.02% | 3.16% | 3.60% |
| n-7 | 2.86% | 3.63% | 3.84% | 4.00% | 3.82% | 4.70% | 4.67% | 4.49% | 4.15% | 3.77% | 3.79% | 3.64% | 4.55% |
| n-6 | 3.40% | 3.71% | 3.90% | 3.97% | 4.96% | 4.53% | 4.67% | 4.45% | 4.02% | 4.41% | 3.89% | 3.91% | 4.22% |
| n-5 | 3.61% | 4.71% | 3.41% | 4.18% | 4.38% | 4.68% | 4.99% | 4.28% | 4.34% | 4.07% | 3.76% | 4.23% | 3.12% |
| n-4 | 3.38% | 4.11% | 4.15% | 3.93% | 4.41% | 4.42% | 5.14% | 4.51% | 4.22% | 4.29% | 3.86% | 4.44% | 3.92% |
| n-3 | 3.62% | 4.97% | 4.40% | 3.51% | 4.36% | 4.61% | 5.11% | 4.21% | 4.26% | 4.55% | 4.03% | 4.17% | 3.46% |
| n-2 | 3.24% | 4.48% | 4.07% | 4.29% | 4.28% | 4.61% | 4.89% | 4.63% | 4.92% | 4.22% | 4.52% | 4.07% | 3.65% |
| n-1 | 3.82% | 4.14% | 4.14% | 3.90% | 4.99% | 5.38% | 5.34% | 5.54% | 5.29% | 4.74% | 4.94% | 4.81% | 4.08% |
| n   | 4.07% | 5.00% | 5.02% | 5.41% | 5.02% | 6.14% | 'x'   |       |       |       |       |       |       |

Table 21: VMS "Herbal" in Currier A[18] – proportion of similar words with an edit distance of two

---

[18] Pages F1v–F25v; F27r–F30v; F32r–F32v; F35r–F38v; F42r–F42v; F44r–F45v; F47r–F47v; F49r–F49v; F51r–F54v; F56r–F56v; F87r–F87v; F90r1–93v; F96r–F96v

# VII. VMS pages with "Herbal" illustrations in Currier B

|     | m-6   | m-5   | m-4   | m-3   | m-2   | m-1   | m     | m+1   | m+2   | m+3   | m+4   | m+5   | m+6   |
|-----|-------|-------|-------|-------|-------|-------|-------|-------|-------|-------|-------|-------|-------|
| n-9 | 0.75% | 0.61% | 0.58% | 0.40% | 0.66% | 0.61% | 0.68% | 0.50% | 0.56% | 0.22% | 0.71% | 0.64% | 0.20% |
| n-8 | 0.33% | 0.34% | 0.51% | 0.54% | 0.66% | 0.56% | 0.49% | 0.37% | 0.51% | 0.97% | 0.52% | 0.16% | 0.40% |
| n-7 | 0.17% | 0.48% | 0.34% | 0.49% | 0.39% | 0.56% | 0.41% | 0.20% | 0.69% | 0.27% | 0.57% | 0.23% | 0.48% |
| n-6 | 0.50% | 0.41% | 0.86% | 0.35% | 0.26% | 0.40% | 0.41% | 0.69% | 0.55% | 0.90% | 0.51% | 0.62% | 0.29% |
| n-5 | 0.50% | 0.14% | 0.40% | 0.73% | 0.52% | 0.67% | 0.70% | 0.56% | 0.85% | 0.47% | 0.55% | 0.52% | 0.09% |
| n-4 | 0.92% | 0.34% | 0.46% | 0.54% | 0.56% | 0.39% | 0.55% | 0.44% | 0.44% | 0.97% | 0.66% | 0.65% | 0.72% |
| n-3 | 0.42% | 0.34% | 0.23% | 0.44% | 0.73% | 0.98% | 0.58% | 0.71% | 0.66% | 0.30% | 0.59% | 0.85% | 0.88% |
| n-2 | 0.17% | 0.69% | 0.63% | 0.83% | 0.69% | 0.66% | 0.83% | 0.66% | 0.73% | 0.64% | 0.63% | 0.48% | 0.60% |
| n-1 | 0.66% | 0.80% | 0.50% | 0.52% | 0.71% | 0.37% | 0.98% | 0.60% | 0.88% | 0.95% | 0.45% | 1.07% | 0.83% |
| n   | 0.65% | 0.98% | 0.91% | 0.85% | 0.70% | 0.78% | 'x'   |       |       |       |       |       |       |

**Table 22: VMS "Herbal" in Currier B[19] – proportion of identical words**

|     | m-6   | m-5   | m-4   | m-3   | m-2   | m-1   | m     | m+1   | m+2   | m+3   | m+4   | m+5   | m+6   |
|-----|-------|-------|-------|-------|-------|-------|-------|-------|-------|-------|-------|-------|-------|
| n-9 | 1.00% | 1.57% | 1.67% | 1.24% | 1.55% | 1.30% | 1.71% | 1.79% | 1.65% | 1.80% | 1.62% | 1.29% | 1.63% |
| n-8 | 1.66% | 0.74% | 1.93% | 1.23% | 1.27% | 1.65% | 1.70% | 1.53% | 1.45% | 1.19% | 0.90% | 1.74% | 1.50% |
| n-7 | 1.92% | 1.56% | 1.60% | 1.23% | 2.14% | 1.71% | 1.78% | 1.87% | 1.56% | 1.33% | 1.83% | 2.08% | 1.74% |
| n-6 | 1.58% | 1.35% | 1.43% | 1.14% | 1.63% | 1.45% | 1.34% | 1.55% | 1.65% | 1.49% | 1.14% | 1.69% | 1.55% |
| n-5 | 1.00% | 1.28% | 1.64% | 1.47% | 1.39% | 1.46% | 1.91% | 1.68% | 1.80% | 1.45% | 1.35% | 0.75% | 1.76% |
| n-4 | 2.17% | 1.84% | 1.88% | 1.67% | 1.69% | 1.69% | 1.72% | 2.11% | 2.22% | 1.37% | 1.08% | 1.66% | 1.35% |
| n-3 | 1.60% | 1.57% | 1.55% | 1.28% | 1.73% | 2.03% | 2.24% | 2.16% | 1.40% | 2.06% | 1.83% | 1.77% | 1.76% |
| n-2 | 2.19% | 1.71% | 2.00% | 2.26% | 1.90% | 1.87% | 1.98% | 1.74% | 1.55% | 1.76% | 1.73% | 1.45% | 1.28% |
| n-1 | 2.39% | 1.54% | 2.45% | 2.09% | 1.87% | 1.95% | 2.06% | 1.74% | 2.13% | 2.34% | 2.29% | 1.81% | 2.57% |
| n   | 1.95% | 2.28% | 2.62% | 2.52% | 2.21% | 2.72% | 'x'   |       |       |       |       |       |       |

**Table 23: VMS "Herbal" in Currier B[19] – proportion of similar words with an edit distance of one**

|     | m-6   | m-5   | m-4   | m-3   | m-2   | m-1   | m     | m+1   | m+2   | m+3   | m+4   | m+5   | m+6   |
|-----|-------|-------|-------|-------|-------|-------|-------|-------|-------|-------|-------|-------|-------|
| n-9 | 3.58% | 3.96% | 3.39% | 3.33% | 3.50% | 4.34% | 4.44% | 3.29% | 3.90% | 3.43% | 4.22% | 2.98% | 4.29% |
| n-8 | 2.74% | 3.11% | 2.84% | 4.03% | 3.07% | 4.27% | 3.81% | 3.31% | 2.57% | 4.00% | 4.65% | 4.04% | 4.30% |
| n-7 | 3.00% | 3.67% | 3.65% | 4.87% | 3.58% | 3.53% | 3.85% | 3.82% | 3.39% | 3.51% | 3.09% | 3.54% | 3.58% |
| n-6 | 3.49% | 3.11% | 3.32% | 4.12% | 3.31% | 3.69% | 3.88% | 3.60% | 4.45% | 2.87% | 3.86% | 4.47% | 4.73% |
| n-5 | 3.68% | 3.38% | 3.52% | 3.82% | 4.39% | 4.02% | 3.64% | 4.04% | 3.55% | 3.11% | 4.25% | 3.35% | 2.87% |
| n-4 | 2.68% | 3.89% | 4.17% | 3.58% | 3.59% | 3.73% | 3.84% | 4.22% | 3.73% | 3.91% | 3.83% | 4.55% | 4.39% |
| n-3 | 3.71% | 4.04% | 3.21% | 4.17% | 3.54% | 4.30% | 4.67% | 3.78% | 3.63% | 3.76% | 4.25% | 4.60% | 4.65% |
| n-2 | 4.05% | 3.98% | 3.37% | 4.07% | 4.05% | 4.24% | 4.17% | 4.14% | 3.87% | 4.17% | 3.85% | 3.51% | 4.69% |
| n-1 | 3.63% | 3.81% | 3.90% | 3.99% | 3.83% | 3.74% | 4.60% | 4.83% | 4.94% | 5.16% | 5.19% | 3.95% | 4.23% |
| n   | 4.71% | 4.03% | 4.87% | 4.09% | 4.96% | 4.66% | 'x'   |       |       |       |       |       |       |

**Table 24: VMS "Herbal" in Currier B[19] – proportion of similar words with an edit distance of two**

---

[19] Pages F26r–F26v; F31r–F31v; F33r–F34v; F39r–F41v; F43r–F43v; F47r–F47v; F50r–F50v; F55r–F55v; F57r; F66v; F94r–F95v1

# VIII. VMS Quire 13 — "Biological" section

|     | m-6   | m-5   | m-4   | m-3   | m-2   | m-1   | m     | m+1   | m+2   | m+3   | m+4   | m+5   | m+6   |
|-----|-------|-------|-------|-------|-------|-------|-------|-------|-------|-------|-------|-------|-------|
| n-9 | 0.96% | 0.61% | 0.65% | 1.08% | 1.02% | 1.30% | 1.22% | 1.33% | 1.13% | 1.15% | 1.14% | 1.07% | 1.04% |
| n-8 | 0.75% | 0.86% | 1.08% | 1.21% | 1.24% | 1.12% | 1.10% | 1.21% | 1.31% | 1.24% | 1.21% | 1.06% | 1.23% |
| n-7 | 1.10% | 0.72% | 1.17% | 1.12% | 1.12% | 1.19% | 1.05% | 1.36% | 1.34% | 1.01% | 1.08% | 0.98% | 0.82% |
| n-6 | 1.11% | 0.83% | 1.14% | 1.10% | 1.25% | 0.93% | 1.36% | 1.17% | 1.32% | 1.21% | 1.13% | 1.29% | 1.01% |
| n-5 | 1.02% | 0.59% | 0.98% | 1.17% | 1.02% | 1.21% | 1.31% | 1.26% | 1.02% | 0.98% | 1.02% | 0.93% | 0.96% |
| n-4 | 0.74% | 1.14% | 1.14% | 1.44% | 1.49% | 1.10% | 1.24% | 1.05% | 1.03% | 0.84% | 1.17% | 0.95% | 0.55% |
| n-3 | 1.04% | 0.96% | 0.87% | 0.98% | 0.95% | 1.30% | 0.99% | 1.13% | 1.22% | 1.21% | 0.80% | 1.09% | 0.90% |
| n-2 | 0.86% | 1.06% | 1.07% | 1.36% | 1.57% | 1.62% | 1.43% | 1.53% | 1.24% | 1.33% | 1.23% | 1.04% | 0.64% |
| n-1 | 1.39% | 1.00% | 1.28% | 1.40% | 1.38% | 1.78% | 1.51% | 1.57% | 1.47% | 1.13% | 1.23% | 1.25% | 1.17% |
| n   | 1.26% | 0.96% | 1.85% | 1.71% | 1.92% | 1.27% | 'x'   |       |       |       |       |       |       |

**Table 25: VMS Quire 13 – proportion of identical words appearing near to the word in writing position 'x'**

|     | m-6   | m-5   | m-4   | m-3   | m-2   | m-1   | m     | m+1   | m+2   | m+3   | m+4   | m+5   | m+6   |
|-----|-------|-------|-------|-------|-------|-------|-------|-------|-------|-------|-------|-------|-------|
| n-9 | 2.72% | 1.81% | 2.29% | 2.83% | 2.60% | 2.66% | 2.76% | 2.27% | 2.33% | 2.20% | 2.21% | 2.31% | 2.14% |
| n-8 | 1.88% | 2.21% | 2.27% | 2.39% | 2.65% | 2.68% | 2.72% | 2.65% | 2.26% | 2.36% | 2.31% | 2.14% | 2.62% |
| n-7 | 2.58% | 2.17% | 2.67% | 2.18% | 2.71% | 2.76% | 2.85% | 2.38% | 2.46% | 2.18% | 2.78% | 2.12% | 2.87% |
| n-6 | 2.30% | 2.75% | 2.52% | 2.30% | 3.03% | 2.74% | 2.51% | 2.50% | 2.83% | 2.82% | 2.73% | 2.61% | 2.54% |
| n-5 | 2.22% | 2.73% | 2.43% | 2.54% | 2.92% | 2.66% | 2.62% | 2.70% | 2.63% | 2.57% | 2.92% | 2.47% | 1.93% |
| n-4 | 2.18% | 2.23% | 2.57% | 2.40% | 2.52% | 3.32% | 2.81% | 2.66% | 2.36% | 2.63% | 2.02% | 2.29% | 2.43% |
| n-3 | 2.11% | 2.32% | 2.43% | 2.53% | 2.89% | 2.97% | 2.58% | 2.83% | 2.53% | 2.97% | 2.48% | 2.50% | 2.58% |
| n-2 | 2.63% | 2.65% | 2.86% | 2.83% | 2.67% | 3.16% | 3.28% | 3.08% | 3.04% | 2.44% | 2.42% | 2.27% | 3.03% |
| n-1 | 2.23% | 2.72% | 2.86% | 2.75% | 2.72% | 3.36% | 3.14% | 2.99% | 3.07% | 3.00% | 3.24% | 3.12% | 3.36% |
| n   | 2.68% | 3.24% | 2.80% | 3.13% | 3.31% | 2.79% | 'x'   |       |       |       |       |       |       |

**Table 26: VMS Quire 13 – proportion of similar words with an edit distance of one**

|     | m-6   | m-5   | m-4   | m-3   | m-2   | m-1   | m     | m+1   | m+2   | m+3   | m+4   | m+5   | m+6   |
|-----|-------|-------|-------|-------|-------|-------|-------|-------|-------|-------|-------|-------|-------|
| n-9 | 3.71% | 3.96% | 4.32% | 4.01% | 5.01% | 4.83% | 4.21% | 4.98% | 4.44% | 4.10% | 4.16% | 3.88% | 3.84% |
| n-8 | 2.96% | 4.97% | 4.04% | 4.74% | 4.31% | 4.96% | 4.39% | 4.60% | 4.48% | 4.04% | 4.44% | 4.60% | 4.31% |
| n-7 | 3.72% | 5.02% | 5.08% | 4.39% | 4.61% | 4.48% | 4.72% | 4.84% | 4.59% | 4.32% | 4.26% | 4.32% | 4.47% |
| n-6 | 4.69% | 4.50% | 4.90% | 4.77% | 4.26% | 4.50% | 4.71% | 4.91% | 4.50% | 4.58% | 4.45% | 4.26% | 4.40% |
| n-5 | 4.13% | 3.91% | 4.99% | 4.65% | 4.24% | 4.87% | 4.39% | 4.87% | 4.65% | 4.33% | 4.53% | 4.06% | 3.74% |
| n-4 | 4.38% | 3.78% | 4.87% | 4.45% | 4.59% | 5.00% | 4.66% | 4.92% | 4.55% | 5.01% | 4.48% | 4.07% | 3.57% |
| n-3 | 4.38% | 4.80% | 4.62% | 4.50% | 4.87% | 4.86% | 4.93% | 4.67% | 4.62% | 4.37% | 4.61% | 4.33% | 4.38% |
| n-2 | 5.10% | 4.13% | 4.71% | 5.04% | 5.11% | 5.38% | 5.48% | 5.24% | 4.51% | 5.35% | 4.31% | 4.33% | 4.60% |
| n-1 | 4.31% | 4.70% | 4.56% | 4.79% | 4.82% | 5.24% | 5.20% | 5.14% | 5.27% | 4.85% | 5.65% | 5.04% | 4.82% |
| n   | 4.71% | 5.13% | 4.90% | 4.78% | 5.59% | 5.07% | 'x'   |       |       |       |       |       |       |

**Table 27: VMS Quire 13 – proportion of similar words with an edit distance of two**

## IX. VMS Quire 20 — "Stars" section

|     | m-6   | m-5   | m-4   | m-3   | m-2   | m-1   | m     | m+1   | m+2   | m+3   | m+4   | m+5   | m+6   |
|-----|-------|-------|-------|-------|-------|-------|-------|-------|-------|-------|-------|-------|-------|
| n-9 | 0.58% | 0.57% | 0.50% | 0.57% | 0.48% | 0.58% | 0.66% | 0.76% | 0.59% | 0.49% | 0.37% | 0.47% | 0.36% |
| n-8 | 0.36% | 0.34% | 0.48% | 0.52% | 0.70% | 0.60% | 0.61% | 0.49% | 0.53% | 0.64% | 0.57% | 0.39% | 0.57% |
| n-7 | 0.41% | 0.34% | 0.57% | 0.55% | 0.60% | 0.52% | 0.63% | 0.64% | 0.54% | 0.55% | 0.67% | 0.49% | 0.48% |
| n-6 | 0.38% | 0.45% | 0.49% | 0.58% | 0.70% | 0.68% | 0.59% | 0.44% | 0.71% | 0.62% | 0.68% | 0.49% | 0.36% |
| n-5 | 0.41% | 0.65% | 0.53% | 0.58% | 0.60% | 0.70% | 0.81% | 0.55% | 0.55% | 0.68% | 0.40% | 0.53% | 0.57% |
| n-4 | 0.60% | 0.33% | 0.47% | 0.52% | 0.54% | 0.83% | 0.72% | 0.50% | 0.58% | 0.58% | 0.52% | 0.47% | 0.26% |
| n-3 | 0.48% | 0.57% | 0.60% | 0.59% | 0.59% | 0.71% | 0.71% | 0.65% | 0.72% | 0.60% | 0.45% | 0.61% | 0.66% |
| n-2 | 0.58% | 0.55% | 0.65% | 0.56% | 0.70% | 0.66% | 0.73% | 0.92% | 0.89% | 0.75% | 0.52% | 0.41% | 0.50% |
| n-1 | 0.55% | 0.53% | 0.48% | 0.52% | 0.73% | 0.62% | 0.76% | 0.72% | 0.73% | 0.66% | 0.79% | 0.50% | 0.48% |
| n   | 0.51% | 0.51% | 0.73% | 0.81% | 0.89% | 0.85% | 'x'   |       |       |       |       |       |       |

**Table 28: VMS Quire 20 – proportion of identical words appearing near to the word in writing position 'x'**

|     | m-6   | m-5   | m-4   | m-3   | m-2   | m-1   | m     | m+1   | m+2   | m+3   | m+4   | m+5   | m+6   |
|-----|-------|-------|-------|-------|-------|-------|-------|-------|-------|-------|-------|-------|-------|
| n-9 | 1.42% | 1.44% | 1.41% | 1.63% | 1.88% | 1.81% | 1.92% | 1.86% | 2.07% | 1.83% | 1.67% | 1.48% | 1.54% |
| n-8 | 1.51% | 1.76% | 1.57% | 1.42% | 1.49% | 1.94% | 1.96% | 1.73% | 1.82% | 1.62% | 1.56% | 1.75% | 1.43% |
| n-7 | 1.70% | 1.93% | 1.93% | 1.64% | 1.79% | 1.99% | 2.23% | 1.97% | 1.58% | 1.79% | 1.69% | 1.97% | 1.51% |
| n-6 | 2.14% | 1.88% | 1.82% | 1.67% | 1.75% | 1.86% | 1.97% | 1.77% | 1.78% | 1.56% | 1.59% | 1.25% | 1.65% |
| n-5 | 1.48% | 1.20% | 1.65% | 1.63% | 1.82% | 1.94% | 1.99% | 2.12% | 1.96% | 1.67% | 1.87% | 1.85% | 1.51% |
| n-4 | 1.46% | 1.75% | 1.87% | 1.29% | 2.01% | 1.91% | 1.89% | 2.22% | 2.13% | 1.81% | 1.83% | 1.57% | 1.58% |
| n-3 | 1.48% | 1.90% | 1.61% | 1.87% | 1.89% | 1.99% | 2.05% | 2.07% | 2.11% | 2.02% | 1.85% | 1.72% | 1.42% |
| n-2 | 1.51% | 1.48% | 1.83% | 1.74% | 2.15% | 2.11% | 2.15% | 1.92% | 1.94% | 1.74% | 1.70% | 1.43% | 1.83% |
| n-1 | 1.45% | 1.65% | 1.85% | 2.13% | 2.19% | 2.19% | 2.54% | 2.22% | 2.21% | 1.78% | 2.00% | 2.02% | 1.87% |
| n   | 2.06% | 1.98% | 2.27% | 2.03% | 2.40% | 2.70% | 'x'   |       |       |       |       |       |       |

**Table 29: VMS Quire 20 – proportion of similar words with an edit distance of one**

|     | m-6   | m-5   | m-4   | m-3   | m-2   | m-1   | m     | m+1   | m+2   | m+3   | m+4   | m+5   | m+6   |
|-----|-------|-------|-------|-------|-------|-------|-------|-------|-------|-------|-------|-------|-------|
| n-9 | 2.98% | 3.38% | 3.53% | 3.77% | 3.65% | 3.66% | 4.04% | 3.68% | 3.69% | 3.37% | 3.73% | 3.63% | 3.58% |
| n-8 | 3.62% | 3.49% | 3.75% | 3.65% | 3.66% | 3.92% | 4.16% | 3.72% | 3.89% | 3.36% | 2.98% | 3.16% | 3.27% |
| n-7 | 3.27% | 3.36% | 3.41% | 3.54% | 3.72% | 3.69% | 4.01% | 3.65% | 3.63% | 3.76% | 3.06% | 3.33% | 3.47% |
| n-6 | 3.46% | 3.62% | 3.68% | 3.54% | 3.97% | 3.72% | 4.12% | 4.01% | 3.74% | 3.81% | 3.44% | 3.78% | 3.34% |
| n-5 | 3.23% | 3.34% | 3.29% | 2.80% | 3.79% | 3.65% | 4.16% | 4.23% | 3.83% | 3.39% | 3.28% | 3.62% | 3.34% |
| n-4 | 3.23% | 3.70% | 3.40% | 3.53% | 3.87% | 4.08% | 3.91% | 3.79% | 3.54% | 3.78% | 3.76% | 3.36% | 3.65% |
| n-3 | 3.28% | 3.53% | 3.80% | 3.60% | 4.06% | 4.08% | 4.07% | 4.08% | 3.85% | 3.83% | 3.53% | 3.54% | 3.18% |
| n-2 | 3.32% | 3.58% | 3.76% | 3.84% | 3.97% | 3.78% | 4.13% | 4.08% | 3.93% | 3.88% | 3.22% | 4.01% | 3.05% |
| n-1 | 3.67% | 3.14% | 3.76% | 3.19% | 3.94% | 4.06% | 4.21% | 3.90% | 3.93% | 4.44% | 4.16% | 3.73% | 3.68% |
| n   | 3.11% | 3.91% | 3.94% | 4.38% | 4.78% | 4.10% | 'x'   |       |       |       |       |       |       |

**Table 30: VMS Quire 20 – proportion of similar words with an edit distance of two**

## X. VMS Quire 20 — "Stars" section without line-initial and line-terminal word

|     | m-6   | m-5   | m-4   | m-3   | m-2   | m-1   | m     | m+1   | m+2   | m+3   | m+4   | m+5   | m+6   |
| --- | ----- | ----- | ----- | ----- | ----- | ----- | ----- | ----- | ----- | ----- | ----- | ----- | ----- |
| n-9 | 0.65% | 0.64% | 0.59% | 0.62% | 0.52% | 0.65% | 0.63% | 0.83% | 0.67% | 0.53% | 0.40% | 0.56% | 0.43% |
| n-8 | 0.46% | 0.41% | 0.52% | 0.56% | 0.72% | 0.63% | 0.62% | 0.56% | 0.57% | 0.75% | 0.64% | 0.46% | 0.71% |
| n-7 | 0.46% | 0.39% | 0.67% | 0.65% | 0.68% | 0.55% | 0.65% | 0.69% | 0.63% | 0.60% | 0.79% | 0.61% | 0.59% |
| n-6 | 0.47% | 0.51% | 0.54% | 0.65% | 0.78% | 0.75% | 0.56% | 0.47% | 0.76% | 0.73% | 0.76% | 0.56% | 0.46% |
| n-5 | 0.49% | 0.73% | 0.60% | 0.66% | 0.68% | 0.75% | 0.78% | 0.58% | 0.61% | 0.80% | 0.44% | 0.56% | 0.68% |
| n-4 | 0.71% | 0.36% | 0.52% | 0.58% | 0.58% | 0.91% | 0.71% | 0.55% | 0.61% | 0.63% | 0.54% | 0.58% | 0.34% |
| n-3 | 0.62% | 0.63% | 0.64% | 0.63% | 0.66% | 0.79% | 0.68% | 0.71% | 0.78% | 0.67% | 0.54% | 0.70% | 0.86% |
| n-2 | 0.74% | 0.61% | 0.77% | 0.64% | 0.77% | 0.75% | 0.74% | 1.04% | 1.01% | 0.86% | 0.60% | 0.51% | 0.49% |
| n-1 | 0.65% | 0.51% | 0.52% | 0.55% | 0.80% | 0.70% | 0.79% | 0.79% | 0.82% | 0.70% | 0.87% | 0.54% | 0.56% |
| n   | 0.66% | 0.58% | 0.88% | 0.86% | 1.00% | 0.92% | 'x'   |       |       |       |       |       |       |

Table 31: VMS Quire 20 – proportion of identical words (line-initial and line-terminal words omitted)

|     | m-6   | m-5   | m-4   | m-3   | m-2   | m-1   | m     | m+1   | m+2   | m+3   | m+4   | m+5   | m+6   |
| --- | ----- | ----- | ----- | ----- | ----- | ----- | ----- | ----- | ----- | ----- | ----- | ----- | ----- |
| n-9 | 1.43% | 1.64% | 1.47% | 1.76% | 2.01% | 1.75% | 2.02% | 2.00% | 2.21% | 1.98% | 1.77% | 1.56% | 1.67% |
| n-8 | 1.67% | 1.95% | 1.63% | 1.57% | 1.67% | 1.97% | 2.02% | 1.79% | 1.90% | 1.72% | 1.68% | 1.86% | 1.50% |
| n-7 | 1.95% | 2.10% | 2.10% | 1.84% | 1.89% | 2.06% | 2.23% | 2.08% | 1.65% | 1.89% | 1.86% | 2.12% | 1.66% |
| n-6 | 2.23% | 1.90% | 2.06% | 1.81% | 1.90% | 1.90% | 1.99% | 1.75% | 1.86% | 1.67% | 1.74% | 1.31% | 1.94% |
| n-5 | 1.63% | 1.24% | 1.70% | 1.70% | 2.05% | 2.00% | 2.10% | 2.13% | 2.10% | 1.69% | 2.03% | 1.87% | 1.57% |
| n-4 | 1.33% | 1.94% | 1.99% | 1.37% | 2.10% | 1.98% | 1.98% | 2.25% | 2.18% | 1.88% | 2.00% | 1.53% | 1.72% |
| n-3 | 1.73% | 1.98% | 1.78% | 2.00% | 1.98% | 2.00% | 2.10% | 2.14% | 2.11% | 2.14% | 2.00% | 1.83% | 1.44% |
| n-2 | 1.74% | 1.69% | 2.09% | 1.74% | 2.20% | 2.25% | 2.29% | 2.01% | 2.03% | 1.88% | 1.89% | 1.53% | 1.99% |
| n-1 | 1.63% | 1.77% | 2.01% | 2.30% | 2.27% | 2.26% | 2.66% | 2.35% | 2.34% | 1.88% | 2.20% | 2.31% | 1.99% |
| n   | 2.37% | 2.10% | 2.46% | 2.07% | 2.58% | 2.85% | 'x'   |       |       |       |       |       |       |

Table 32: VMS Quire 20 – proportion of similar words with an edit distance of one (line-initial and line-terminal words omitted)

|     | m-6   | m-5   | m-4   | m-3   | m-2   | m-1   | m     | m+1   | m+2   | m+3   | m+4   | m+5   | m+6   |
| --- | ----- | ----- | ----- | ----- | ----- | ----- | ----- | ----- | ----- | ----- | ----- | ----- | ----- |
| n-9 | 3.29% | 3.69% | 3.71% | 3.94% | 3.86% | 3.70% | 4.09% | 3.81% | 3.95% | 3.66% | 3.97% | 3.85% | 4.02% |
| n-8 | 3.96% | 3.60% | 3.69% | 3.99% | 3.71% | 4.12% | 4.27% | 3.91% | 4.17% | 3.53% | 3.12% | 3.34% | 3.59% |
| n-7 | 3.28% | 3.51% | 3.61% | 3.71% | 3.98% | 3.91% | 4.11% | 3.76% | 3.86% | 3.99% | 3.29% | 3.70% | 3.76% |
| n-6 | 3.66% | 3.81% | 3.89% | 3.70% | 4.16% | 3.72% | 4.21% | 4.15% | 3.90% | 3.93% | 3.66% | 3.99% | 3.53% |
| n-5 | 3.45% | 3.76% | 3.66% | 3.10% | 4.04% | 3.77% | 4.16% | 4.52% | 4.19% | 3.50% | 3.55% | 3.81% | 3.66% |
| n-4 | 3.45% | 3.83% | 3.58% | 3.69% | 4.11% | 4.23% | 3.96% | 3.93% | 3.71% | 4.05% | 3.97% | 3.55% | 4.12% |
| n-3 | 3.46% | 3.78% | 3.96% | 3.75% | 4.29% | 4.17% | 4.17% | 4.19% | 4.11% | 4.02% | 3.67% | 3.76% | 3.27% |
| n-2 | 3.38% | 3.69% | 4.07% | 3.99% | 4.19% | 3.86% | 4.30% | 4.26% | 4.17% | 4.16% | 3.40% | 4.24% | 3.34% |
| n-1 | 3.84% | 3.37% | 3.99% | 3.22% | 4.13% | 4.16% | 4.22% | 3.97% | 4.14% | 4.66% | 4.56% | 4.10% | 3.76% |
| n   | 3.18% | 4.30% | 4.10% | 4.59% | 4.93% | 4.21% | 'x'   |       |       |       |       |       |       |

Table 33: VMS Quire 20 – proportion of similar words with an edit distance of two (line-initial and line-terminal words omitted)